\documentclass{article}
\usepackage{spconf,amsmath,epsfig}
\usepackage{times}
\usepackage{graphicx}
\usepackage{amssymb}
\usepackage{xcolor}
\usepackage{url}
\usepackage{epstopdf}
\usepackage{xspace}
\usepackage{caption}
\usepackage{subcaption}
\usepackage{tikz}
\usepackage[absolute]{textpos}

\newcommand*{\eg}{e.g.\@\xspace}

\TPMargin{5pt}

\newcommand{\copyrightstatement}{
    \begin{textblock}{0.8}(0.1,0.93)    
         \noindent
         \footnotesize
         \copyright  2015 IEEE. Personal use of this material is permitted. Permission from IEEE must be obtained for all other uses, in any current or future media, including reprinting/republishing this material for advertising or promotional purposes, creating new
collective works, for resale or redistribution to servers or lists, or reuse of any copyrighted component of this work in other works. DOI: 10.1109/ICME.2015.7177502
    \end{textblock}
}

\begin{document}\sloppy
\copyrightstatement
\def\x{{\mathbf x}}
\def\L{{\cal L}}

\title{Relative Learning from Web Images for Content-adaptive Enhancement}
%
\name{Parag Shridhar Chandakkar, Qiongjie Tian and Baoxin Li}
\address{School of Computing, Informatics and Decision Systems Engineering, Arizona State University \\
{\tt\small \{pchandak,qtian5,baoxin.li\}@asu.edu}}
%
%
%

\maketitle

\begin{abstract}
Personalized and content-adaptive image enhancement can find many applications in the age of social media and mobile computing. This paper presents a relative-learning-based approach, which, unlike previous methods, does not require matching original and enhanced images for training. This allows the use of massive online photo collections to train a ranking model for improved enhancement. We first propose a multi-level ranking model, which is learned from only relatively-labeled inputs that are automatically crawled. Then we design a novel parameter sampling scheme under this model to generate the desired enhancement parameters for a new image. For evaluation, we first verify the effectiveness and the generalization abilities of our approach, using images that have been enhanced/labeled by experts. Then we carry out subjective tests, which show that users prefer images enhanced by our approach over other existing methods.
\end{abstract}
\section{Introduction}
\label{sec:intro}

In today's age of social media, it is becoming more important to capture good-looking photos. Due to the outreach of social media sites, the photos get spread around quickly. It is common to retouch the photo after capturing to improve its appearance. The photo-retouching tools have made significant progress in recent years. There exist sophisticated tools such as Adobe Photoshop as well as one-touch enhancement tools such as Picasa, Windows Live Gallery and Apple's auto-enhance. However, one-touch enhancement tools neither offer personalization nor content-based image enhancement. For example, an indoor image may need a different style of enhancement than an outdoor image. Adobe Photoshop offers large variety of enhancement operations but can be complex and time-consuming for an amateur photographer. This underlines the need for better and automated image enhancement tools. Enhancement operations are performed on various aspects of an image, e.g. saturation, contrast, brightness, sharpness, etc. Hence the space of possible combinations of enhancement parameters is huge. This work focuses on content-based image enhancement by using content-similar high-quality images as reference.

Training-based methods have recently been explored for image enhancement, where pairs consisting of a low-quality image and its enhanced counterpart are used for training \cite{learningToRank,caicedo2011collaborative,hwang2012context,kang2010personalization}. The enhancements are done by expert users. Such a training set allows learning of a regression/ranking function which maps the input feature to the optimal enhancement parameters. For a regression function, it learns a mapping between the input parameters (could be pixel values) to the parameters in the corresponding enhanced image. The ranking function assigns a score to each feature vector. The enhancement parameter which gets the highest score is selected as the best enhanced version of the input image. However, such schemes do not scale well, owing to the need of expert-enhanced training images. Per our knowledge, the largest such publicly available training database is MIT-Adobe 5K \cite{bychkovsky2011learning}, consisting of 5 enhanced versions per image and 5000 images. In reality, we have millions of high-quality images available on the Web which, if properly utilized, can improve the performance significantly. Further, since the previous approaches need low-quality and its enhanced counterparts, it is difficult to customize the system according to the individual's preferences. Our approach can handle a non-corresponding pair of a low-quality image and a high-quality reference image. We can possibly retrieve popular images from a user's Flickr/Instagram account to customize the enhancement preferences. To the best of our knowledge, only our approach considers both of the above aspects simultaneously.

It is a challenge to find optimal enhancement parameters with non-corresponding pair of input and output images. The visual features are not corresponding to each other to build a regression or a simple ranking function. Usually, the optimal parameters of low-quality image are explored near the parameter space of its enhanced counterpart. In this case, the search for possible space of enhancement parameters is extremely difficult due to non-correspondence of input and output. To remedy this, we define a novel parameter sampling scheme and a multi-level ranking model which uses simple visual features along with derived image parameter features (such as brightness, contrast and saturation). We build a multi-level ranking relation from the partial ordering available between the visual feature and parameter vectors of low-quality input and high-quality reference images. A learning-to-rank approach has already been proposed in \cite{learningToRank}. Unlike us, they need the corresponding pairs of input and output images generated by an expert user along with a record of the intermediate enhancement steps. This limits the possible applications of their approach as discussed before. We show superiority of our method over one-touch enhancement tools and the state-of-art ranking-based enhancement approach \cite{learningToRank}.


\begin{figure}[!t]
\centering
\includegraphics[scale=0.05]{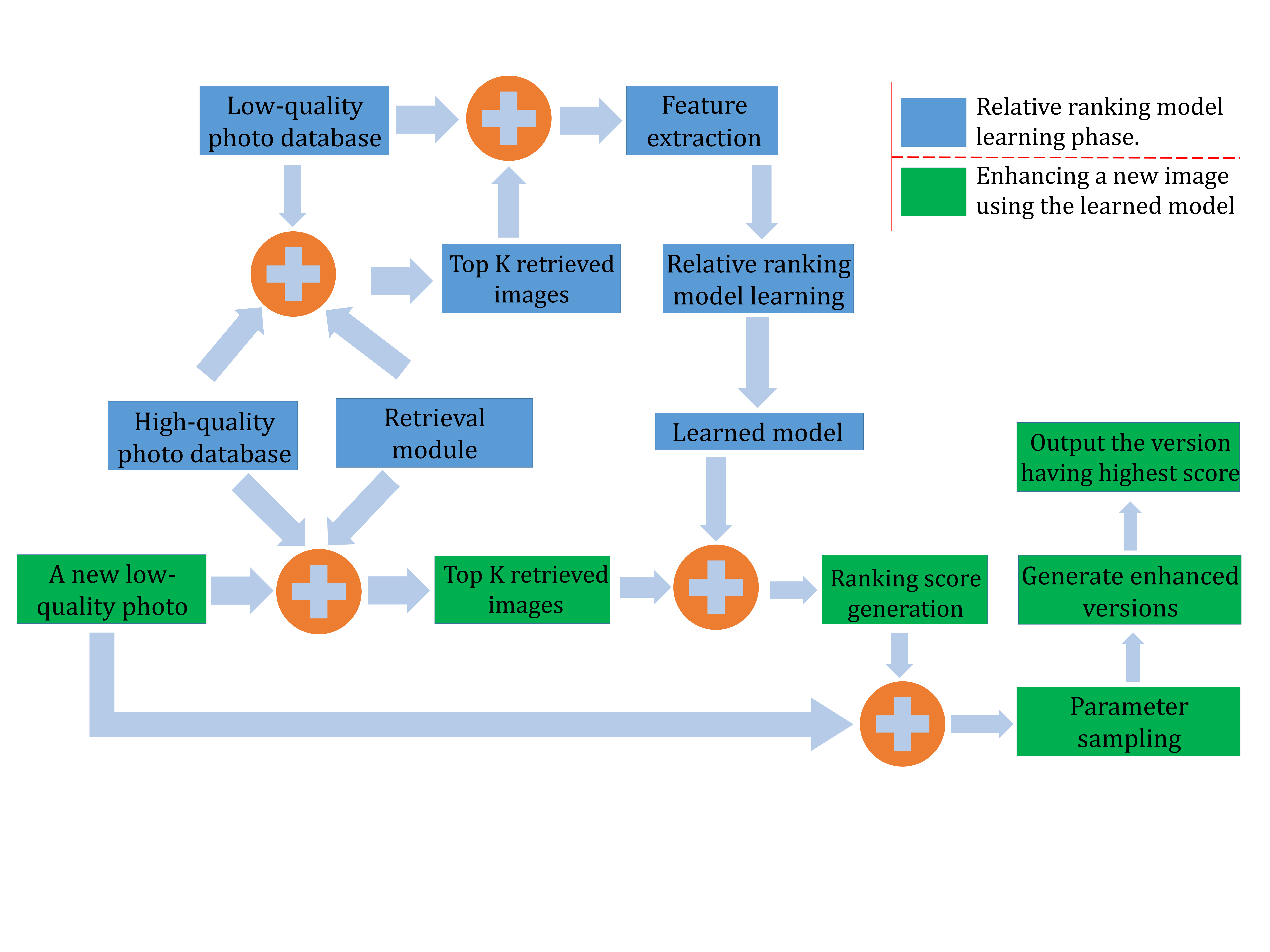}
\caption{Overview of our image enhancement algorithm.}
\label{fig:frameworkFlow}
\end{figure}	

\section{Related Works}

Purely image-processing-based and learning-based approaches have been developed for enhancing an image. In this section, we mainly focus on learning-based approaches. Common faces across the images have been used in \cite{joshi2010personal} for personal photo enhancement. They built a system to detect examples of good and bad faces. Then the good faces were used for enhancing the bad ones. However, their approach lacks generality. A novel tone-operator was proposed in \cite{reinhard2002photographic} to solve the tone reproduction problem. In \cite{caicedo2011collaborative}, a preliminary solution was proposed to enhance an image according to a user's profile. The result shows that the users' preference can be classified into different groups and also improved enhanced results can be obtained by enhancing images according to users' choices. Local image enhancement was performed in \cite{hwang2012context} by using local scene descriptors along with context. For different scenes in the input image, similar image pairs are retrieved. Then for each pixel in the input image, a set of pixels were retrieved from the data-set and used to improve the given pixel. Then Gaussian random fields are used to maintain spatial smoothness in the enhanced image and improve the perceptual quality of the image. MIT and Adobe collaborated to generated a large reference data-set which has $5000$ input images and each has five enhanced versions, created by five experts \cite{bychkovsky2011learning}. Using this database, they apply supervised learning to predict a user's adjustment and the preference for a new user. They also analyze the difference in users' preferences. In \cite{kang2010personalization}, the user preference was modeled based on the image database. Users have to enhance some images to effectively train the model. The learned model was then applied to obtain multiple enhanced versions for an image, according to the user's preferences. Image enhancement based on content and scene semantics was done in \cite{kaufman2012content}. Regions containing different objects were first detected. Then customized enhancement operators were applied in these regions. They concentrated more on personal photo collections and their content detectors were limited to some objects such as people, sky etc. The intermediate enhancement steps carried out by an expert were recorded and used to train a ranking model in \cite{learningToRank}. They generated multiple enhanced versions by sampling in the enhancement parameter space. The enhanced version which obtained the highest ranking score was selected as the final output. However, they require a lot of work from the expert which is undesirable due to aforementioned reasons.

\section{Proposed Approach}

Our approach enhances an image depending on its content and color composition. Fig. \ref{fig:frameworkFlow} shows the flow of our algorithm. We first perform CBIR to retrieve high-quality images similar to the low-quality input image. We then create a ranking between the input image and the retrieved high-quality images using extracted image features. To overcome the shortcoming of not having one-to-one correspondence between the query and the retrieved high-quality images, we introduce an additional level in the ranking problem (detailed later) to create a three-level ranking problem. For a new low-quality image, we generate multiple enhanced versions of the input image by using a novel parammeter sampling scheme. We select the version which gets highest rank from the model. The individual steps of our approach are detailed below.

\subsection{Data Collection and Processing} \label{sec:dataCollProc}

A good collection of high-quality images is essential for our approach. We choose high-quality images from the database published in \cite{datta2008}. It consists of photos from \textit{DpChallenge.com} and \textit{Photo.net} among others. We select top $10\%$ photos which have more than 10 ratings. We then remove grayscale photos. DpChallenge.com has photos based on $66$ themes ranging from ``animals'', ``food'' to ``portrait'', ``water'' etc. Our database contains $1,822$ and $9,467$ photos from Photo.net and DpChallenge.com respectively. We use bottom $10\%$ rated photos on DpChallenge.com as low-quality photos. We also select photos from Flickr taken by old camera-phones such as iPhone 3G, Samsung Galaxy II for low-quality photos.

We use an open-source CBIR engine called LIRE \cite{lire} to retrieve $K$ (=$100$) high-quality images for each low-quality image. In the CBIR framework, we use opponent histogram and auto color-correlogram features for color, PHOG descriptor and JPEG coefficient histogram to represent shape and image quality respectively. Before retrieval, we introduce a simple but effective step of auto-enhancement. It increases the contrast of an image by saturating $1\%$ of its data at low and high intensities. Low-quality images usually have less contrast. Some of those images look either hazy or dark. Therefore, the retrieved images do not correspond to the true colors in the low-quality image. This is avoided by first auto-enhancing each low-quality image before retrieval.

\subsection{Relative Ranking Model Learning}

We formulate the problem of enhancing a low-quality image using other retrieved high-quality images as a multi-level ranking problem. We now explain the general formulation of our ranking approach. It is followed by the proposed multi-level ranking and the formation of feature vectors.


Suppose $Q$ and $D$ are the sets of query images and all high-quality images respectively. The retrieved images of the given image, $I_i \in Q$, are denoted as $R_i \subseteq D$. We assume that images in $R_i$ are of better quality than $I_i$. We construct relative ranking pairs, say $(I_i, R_{i,j})$ where $R_{i,j}$ is the $j^{th}$ image in $R_i$. These pairs are now used for building a simple two-level ranking model. We use visual features along with the parameter feature vectors in our ranking model, while \cite{learningToRank} uses only visual features. The parameters help us to capture hidden characteristics of an image. For example, the contrast of an image depends on the spatial arrangement of the various colors. Saturation represents the purity of the color in a different dimension. The combination of features and parameters helps compare seemingly unrelated input and output images in a similar feature space. Let us denote the visual features extracted for $I_i$ as $V_i$ and the feature vector for the $k^{th}$ parameter as $P_i^k$. Then the ranking model is given as follows:

\begin{equation}
\begin{aligned} \label{eqn:rank_model_1}
\mbox{$\min$ \quad} & \frac{1}{2} \omega^2 + \sum_{i,j}\xi_{i,j} \\
\mbox{s.t. \ \quad} & \omega^Tf_i \leq \omega^T g_{i,j} - \xi_{i,j} \quad \forall \ i,j \\
& \xi_{i,j} \geq 0
\end{aligned}
\end{equation}
where $f_i = [V_i, P_i^1, \hdots, P_i^k]$ and $g_{i,j} = [V_{R_{i,j}}, P_{R_{i,j}}^1, \hdots, \allowbreak  P_{R_{i,j}}^k]$. 

\vspace{2pt}
The above ranking model concatenates all the feature and parameter vectors to get a score for an image. However, assume a situation where we have an image with a slightly higher level of saturation and brightness but with less contrast. Since the above ranking model considers all the elements together, the image may obtain a high score. Thus we propose a modified ranking model in which we concatenate visual feature and \textit{one} parameter at a time.
\vspace{-2pt}
\begin{equation}
\begin{aligned} \label{eqn:rank_model_2}
\hspace{-10pt}\mbox{$\min$ \quad} & \frac{1}{2} \omega^2 + \sum_{i,j}\xi_{i,j} \\
\mbox{s.t. \ \quad} & \omega^Tf_i^n \leq \omega^T g_{i,j}^n - \xi_{i,j}^n \mbox{  } \quad \forall \ i,j,n \ \text{and} \ n=\{1,\hdots,k\}\\
& \xi_{i,j}^n \geq 0
\end{aligned}
\end{equation}
where $f_i^n = [V_{I_i}, P_i^n]$ and $g_{i,j}^n = [V_{R_{i,j}}, P_{R_{i,j}}^n]$. 	


\vspace{2pt}
In Equation \ref{eqn:rank_model_2}, we would get as many constraints as the cardinality of $n$. In other words, it will be equal to the total number of parameters. The final score depends on the combination of visual feature and \textit{each} parameter. An image gets a high score only if all of its parameters are in balanced amounts. Moreover, the visual feature stays common in all of the inequality constraints since the parameters are dependent and changing one of them affects all of them. A common visual feature ensures that only a balanced feature-parameter combination defines a high-quality image. All parameter feature vectors, $[P_i^1,\hdots,P_i^k]$, are required to have the same length in order to construct such a model.

\vspace{-6pt}
\subsubsection{Multi-level ranking}
We enhance the low-quality images using the high-quality ones stored on photo-sharing websites. Unlike in \cite{learningToRank}, we neither possess corresponding pairs of original and enhanced images nor the record of intermediate steps carried out by an expert during an enhancement process. The external high-quality images have high contrast, brightness and saturation. Therefore, the ranking model generated using Equation \ref{eqn:rank_model_1} or \ref{eqn:rank_model_2} has only learned that high values can generate a high-quality photograph. The model lacks the knowledge that extremely less or high values of parameters (\eg brightness, saturation and contrast) can degrade the image quality significantly.

To incorporate this knowledge, we propose multi-level ranking. We manually vary brightness, contrast and saturation for 20 images in our database (one parameter at a time) till their quality degrades significantly. The variation happens on both the extremes. For the rest of the images, we automatically vary the parameters to generate $8$ degraded versions for each low-quality image. The amount of variation in the parameters is determined empirically using these $20$ images.

Let us denote the $m^{th}$ corresponding degraded image for a low-quality image, $I_i$, by $B_i^m$. The visual feature vector and the $k^{th}$ parameter vector of \vspace{1.5pt}$B_i^m$ is denoted by $V_{B_i^m}^k$ and $P_{B_i^m}^k$ respectively. We use $B_i^m$, $I_i$ and $R_{i,j}$ to build a three-level ranking model such that $B_i^m < I_i < R_{i,j} \quad \forall \quad \{i,j,m\}$. The ranking model can be formulated as follows:
\vspace{-1pt}
\begin{equation}
\begin{aligned} \label{eqn:rank_model_3}
\hspace{-50pt}\mbox{$\min$ \quad} & \frac{1}{2} \omega^2 + \sum_{i,j}\xi_{i,j} + \sum_{i,m}\xi'_{i,m} \\
\mbox{s.t. \ \quad} & \omega^T h_i^m - \xi'_{i,m} \leq \omega^Tf_i \leq \omega^T g_{i,j} - \xi_{i,j} \\
& \xi_{i,j} \geq 0, \ \xi'_{i,m} \geq 0 \quad \forall \ i,j,m \mbox{  }, \ m=\{1,\hdots,8\}
\end{aligned}
\end{equation}
where $f_i=[V_i, P_i^1, \hdots, P_i^k]$ and $g_{i,j} = [V_{R_{i,j}}, P_{R_{i,j}}^1, \hdots, \allowbreak P_{R_{i,j}}^k]$ and $h_i^m=[V_{B_i^m}, P_{B_i^m}^1, \hdots, P_{B_i^m}^k]$.
\vspace{1.5pt}
The approach using above ranking model is termed as \textit{approach-3176} since the feature space generated using this approach is $3176$-D . Similarly, Equation \ref{eqn:rank_model_2} can be converted into a three-level ranking model and we call that approach as \textit{approach-2744}. All the ranking formulations are solved using ranking-SVM \cite{joachims2006training}.

\subsubsection{Feature and parameter vectors for ranking}

The entire feature vector includes a 2600-D HSV histogram (visual feature) and four 144-D parameter vectors representing brightness, contrast, saturation and sharpness.

\vspace{3pt}
\noindent \textbf{HSV histogram}: We divide saturation and value uniformly into 10-bins each. Hue is divided non-uniformly into 27 bins, based on the distance between the hues in the CIELAB color space. That distance is given by the CIEDE2000 metric which considers perceptual uniformity. The reader is pointed to \cite{sharma_ciede2000} for the representation of the formula and its implementation details. We form a 2-D grid where saturation and brightness varies from $0$ to $1$ in a grid and hue varies across different grids, in steps of $5$. We measure the CIELAB distance between the corresponding points on two grids. The maximum distance between two grids gives us a rough measure of the distance between all possible shades of these two hues. Thus,

\begin{equation}
d_{mn}=\max_{ij} \Delta E(G_{H_m}(i,j),G_{H_n}(i,j))
\end{equation}

\vspace{3pt}
\noindent where $G_{H_m}$ and $G_{H_n}$ are the grids for the $m^{th}$ and $n^{th}$ hue respectively. $\Delta E$ is the CIEDE2000 color difference metric. $d_{mn}$ provides the largest possible distance between all shades of hue $m$ and $n$. Our aim is to make a separate bin for a hue which is significantly different from the previous hue. To achieve this, we keep $m$ constant and vary $n$ till $d$ crosses a pre-determined threshold (=$7$). Once this threshold has been reached, we assign the value of $n$ to $m$ and repeat the process till $m$ reaches $360$. We obtain $27$ bins for hue. Due to the proposed binning, the HSV histogram captures more details in an image by using relatively less number of bins. Separation between two hues ranges from $6^{\circ}$ to $40^{\circ}$ in this binning.

\vspace{3pt}
\noindent \textbf{Contrast}: We propose to measure local RMS contrast. The RMS contrast of an $M \times N$ image $I$ is defined as the standard deviation of pixel intensities as follows,

\begin{equation}
C_{RMS}=\sqrt{\frac{1}{MN} \sum_{i=0}^{N-1} \sum_{j=0}^{M-1}(I-\bar{I_{ij}})}
\end{equation}

\noindent We first divide the image into $12 \times 12$ grid to capture the local characteristics of an image. Each grid is subdivided into blocks of $8 \times 8$ pixels. We measure the RMS contrast of these blocks and average the contrast values inside a grid. This gives a 144-D vector ($12 \times 12 = 144$) for each image.

\vspace{3pt}
\noindent \textbf{Brightness and Saturation}: We divide the image into $3 \times 3$ grid. For each grid, we calculate a 16-bin histogram of $V$ and $S$-channel. This gives us two 144-D ($16 \times 9=144$) histograms as brightness and saturation features.

\vspace{3pt}
\noindent \textbf{Sharpness}: We adopt the approach mentioned in \cite{photoQuality} to measure sharpness. Instead of calculating the sharpness metric for the entire image, we divide the image into $3 \times 3$ grid and for each grid, we calculate its sharpness as the ratio of area of high-frequency components to the total area of the grid,

\begin{equation}
m_{sharp}=\frac{||H||}{||I||},
\end{equation}

\noindent where $||H||=\{(u,v) \vert \hspace{3pt} |F(u,v)| > \theta\}$ and $F$ is the FFT of image $I$. $\theta$ is a pre-defined threshold. By varying $\theta$, one can decide the sharpness level of an image, at which the metric should start responding. For example, $m_{sharp}$ will produce large values even for a blurred image when $\theta$ is kept small and vice-versa. We define $\theta$ as a monotonically-increasing 16-dimensional vector. $\theta$ increases on a log-scale from $0$ to $1$. The $i^{th}$ bin of the sharpness histogram is defined as,

\begin{equation}
m_{{sharp}_i}=\frac{||H_i||}{||I||} \hspace{15pt} \text{for} \hspace{4pt} i=[1,...,16],
\end{equation}

\noindent where $||H_i||=\{(u,v) \vert \hspace{3pt} |F(u,v)| > \theta(i)\}$. $m_{sharp}$ is calculated for each grid to produce a 144-D sharpness feature.

\subsection{Algorithm for enhancing a new image}

The relative ranking model is now capable of assigning a score to an enhanced version of an image depending on its color composition and content. However,  generating these enhanced versions is not straightforward due to the large number of possible enhancement parameter combinations. We employ a novel parameter sampling scheme based on the ranking scores to reduce the search space as follows.

We retrieve $100$ most similar images to the new image by using the CBIR module defined in Section \ref{sec:dataCollProc}. We then calculate scalar values for saturation, brightness and contrast of the retrieved images, denoted by $s_R,b_R$ and $c_R$ respectively. We also define the terms $s_N,b_N$ and $c_N$ for the new image equivalently. The average saturation and brightness is calculated as the mean of $S$ and $V$ channel of an image in $HSV$ color space, respectively. The average contrast is the standard deviation of the $RGB$ pixel values. Next step is to calculate the ranking scores of those retrieved images by multiplying the learned model $w$ with the feature vectors of all the retrieved images. The ranking score for the $i^{th}$ retrieved image is denoted by $scr_i$. The procedure of calculating $scr$ for the approach-$3176$ and $2744$ is slightly different as follows:

\noindent
\vspace{-6pt}
\begin{equation}
scr_{3176}=w^T*f, \quad scr_{2744}=\frac{1}{k} \ \sum_{j=1}^{k}{w^T*f^j}
\end{equation}

\noindent
where $f$ and $k$ are the feature vector of a retrieved image and number of parameters respectively. Note that since the features for both the above approaches are different, the learned models have a different structure ($3176$-D vs. $2744$-D).

We non-uniformly sample around the values of $s_R,b_R$ and $c_R$ using the obtained ranking scores. Dense (sparse) sampling is performed around those values of $s_R,b_R$ and $c_R$ for which we have obtained high (low) scores. A high score indicates that the image is visually appealing and similar to the new image. After parameter sampling, we try to steer $s_N,b_N$ and $c_N$ towards the sampled values. However, the low and high-quality images are not counterparts of each other and hence we stop steering $s_N, b_N$ and $c_N$ towards the sampled values if the percentage change is more than a certain threshold. It is determined empirically to be $\pm20\%$ for average brightness and saturation and $\pm4\%$ for average contrast. Using this procedure, we generate anywhere from 150 to 250 enhanced versions of an image. Ranking scores are calculated for all these enhanced versions as well as the original one. The image with the highest score is presented to the user.

\section{Experiments and Results}

We carried out two-fold assessment of our algorithm. Firstly, as a verification step, we evaluated our ranking model on manually enhanced and degraded images. We selected $500$ images from MIT-Adobe FiveK dataset \cite{bychkovsky2011learning}. Each image in the data-set has been enhanced by five experts. We also generated $8$ degraded versions for every image. We calculated ranking scores for original images, its enhanced and degraded counterparts. For $90.19\%$ ($451/500$) images, at least one enhanced version got a higher score than the original image. For $100\%$ of the images, at least one bad version obtained less score than the input image. For $80.27\%$ ($401/500$) of the images, at least 7 out of 8 bad versions got a lower score than the input image.

For a more robust performance assessment, we carried out a subjective evaluation test. To this end, we formed a testing set with $127$ images which is disjoint with the training set. Our low-quality image data-set contains $77$ images. The remaining $50$ images were selected at random from the 94 image database provided to us by the authors of \cite{learningToRank}. Our subjective evaluation test involved $33$ users. Each user was shown a pair of images and was asked to choose the better photograph. In case a user preferred both photos equally, a third option of voting to both of them was made available. There was no time-limit and users could take breaks in between if they were fatigued. The tests were done on the same type of monitor and the lighting conditions as well as the sitting arrangements were identical for all the tests. The order between different pairs and between the images of a pair was kept random. No indication regarding the type of enhancement method used was given to the user.

The subjective evaluation test consisted of five sessions. Users were asked to compare approach-2744 and 3176 to Picasa in the first two sessions. In the next two sessions, we enhanced the $50$ images using these two approaches and asked users to compare them with the approach of \cite{learningToRank}. Finally, we skipped the auto-enhancing step mentioned in the Section \ref{sec:dataCollProc} and also employed a simple two-level ranking (instead of the three-level ranking) as described in Equation \ref{eqn:rank_model_1} to formulate a new but presumably inferior approach. It is now used to enhance $25/77$ images, chosen at random. In the fifth session, users compared these $25$ images with Picasa. The last session aims to explain the need of multi-level ranking and pre-processing step of auto-enhancement before retrieval. In total, each user compared $279$ ($=77*2$ + $50*2$ + $25$) pairs of images. Each user took anywhere from $30$-$45$ minutes to complete the test.

\begin{figure}[!t]
\centering
\includegraphics[scale=0.55]{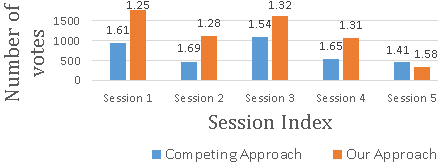}
\caption{Average result of our subjective evaluation study.}
\label{fig:allSessions}
\end{figure}


Fig. \ref{fig:allSessions} shows the cumulative votes given to each approach in every session.  The votes are accumulated from all users over all the images. The method which was selected more number of times in a session was favored by the users for that session. Users consistently pick photos enhanced by our approach over Picasa and the approach of \cite{learningToRank}. Interestingly, the results of the fifth session favor Picasa, implying the need for multi-level ranking. The numbers above each bar represent the average rank obtained for each approach, lower rank implying better quality for the image. It is calculated as the average of all the vote ratings (from all users) to that approach.

\begin{figure}[!t]
\centering
\includegraphics[width=0.42\textwidth,height=250pt]{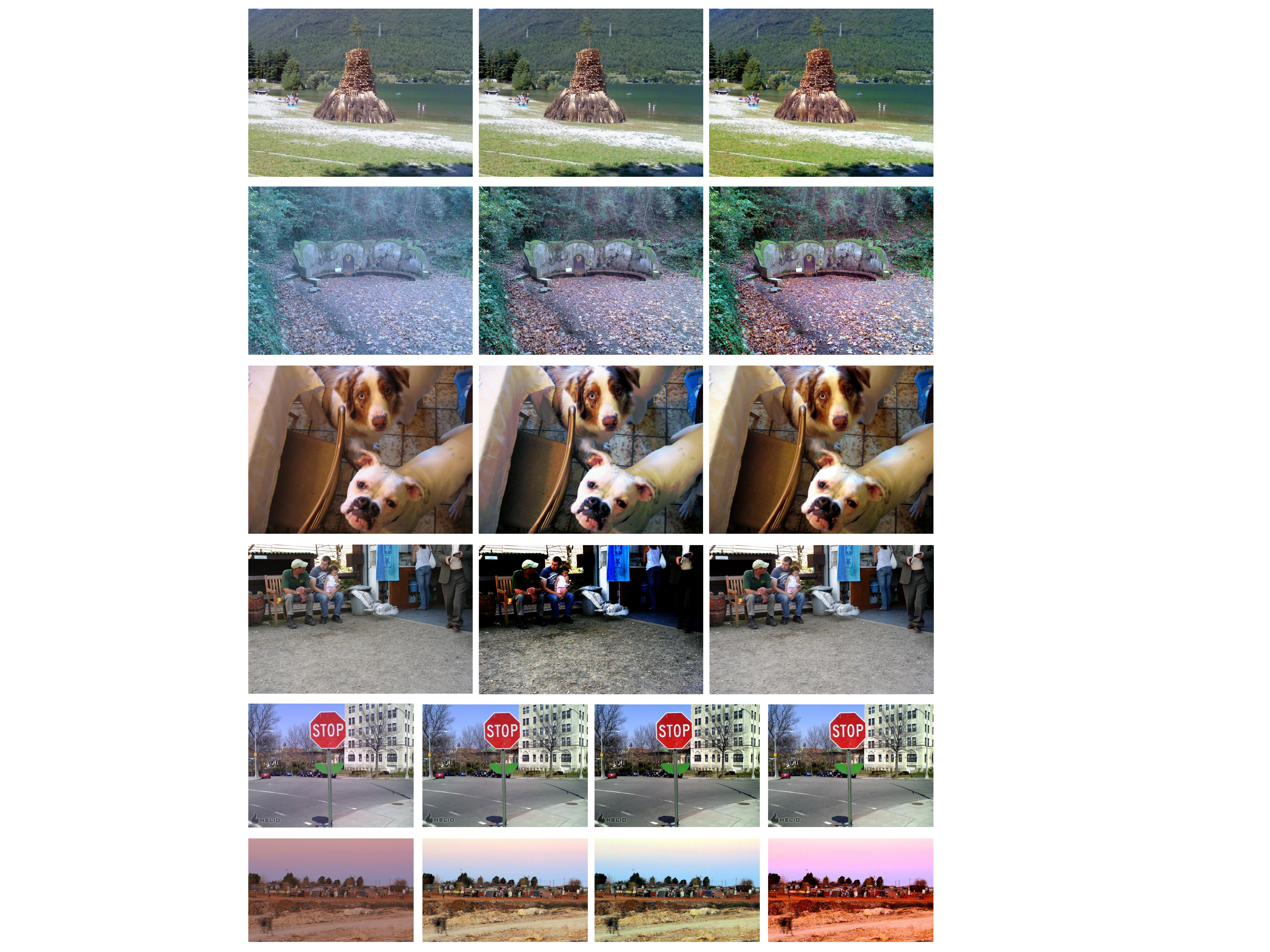}
\caption{Comparison of image enhancement methods. \textbf{Top two rows, our data:} left: original image, middle: enhancement by Picasa, right: approach-2744. \textbf{Row 3 and 4, data of \cite{learningToRank}:} left:original image, middle: enhancement result of \cite{learningToRank}, right: approach-2744. \textbf{Row 5, our data:} From L to R, 1. original image, 2. enhancement by Picasa, 3. approach-3176, 4. approach-2744. \textbf{Last row, our data:} From L to R, 1. original image, 2. enhancement by Picasa 3. multi-level ranking, approach-3176 4. two-level ranking, approach-3176.}
\label{fig:imageEnhancementResult}
\end{figure}

Fig. \ref{fig:imageEnhancementResult} shows example results from session $1,2$ (row 1, 2, 5), $3$ (row 3, 4) and $5$ (row 6). Similar to \cite{learningToRank}, even we observed that Picasa is conservative while enhancing photos. Picasa concentrates on conservatively adjusting the brightness and contrast. However, images in the second row illustrate that the conservative adjustment does not suit all the images. Our approach converges on better enhancement parameters by performing content-adaptive enhancement. For example, outdoor images may need significant changes in their composition while quality of indoor images may be harmed by doing so. On the other hand, the approach of \cite{learningToRank} significantly changes the original photo as shown in the middle two images in row 3 and 4. However, in this process, sometimes large variation happens in the amount of contrast and saturation, which harms the quality of the photo. We avoid this by introducing multi-level ranking and the novel parameter-sampling strategy. Approach of \cite{learningToRank} enhances the original photo based on the parameters of $K$-nearest high-quality images (in terms of $L2$ distance) alone. We sample our parameters based on $K$-nearest images and more importantly using the ranking score. We also set an upper limit to the amount of variation in the enhancement parameters. Thus our parameter sampling takes into account the color composition, content and quality of the retrieved images. The importance of these steps is clearly seen in the images in the last row. The rightmost image is over-saturated since the model lacks any knowledge about the bad effects of setting extreme values for image parameters.

Our algorithm prefers more significant adjustments than Picasa. The reason for such a preference stems from our training database. We have total $11,289$ images from \textit{DPChallenge.com} and \textit{Photo.net}. Most of the images are vibrant in colors, with high-contrast and saturation. Though many participants in our study tend to choose high-contrast and saturation photos, it is possible that some users prefer washed-out or dark photos. Such personalized preferences can be learned by our model by training on users' Flickr or Instagram feed.

\section{Conclusion} \label{sec:conclusion}

We presented a novel learning-based framework for image enhancement. It uses CBIR to perform content-adaptive enhancement of low-quality images. A multi-level relative ranking model is trained with the help of high-quality images on the web. We show that instead of concatenating all features, considering them pairwise in the ranking model creates better enhanced images with all of its parameters in balanced amounts. We propose a novel parameter sampling scheme to reduce the huge search space and converge onto better enhancement parameters. We verified the effectiveness of our framework by checking its performance on MIT-Adobe FiveK dataset. For a more robust comparison, we carry out subjective evaluation tests and show that users prefer photos enhanced by our framework over others. Our framework offers scalablity and personalization since it directly uses high-quality image databases from the web.
	
\textbf{Acknowledgement:} The work was supported in part by a grant (1135616) from the National Science Foundation. Any opinions expressed in this material are those of the authors and do not necessarily reflect the views of the NSF.

\bibliographystyle{IEEEbib}
\bibliography{egbib}

\end{document}